\pdfoutput=1
\documentclass{article}
\PassOptionsToPackage{numbers,compress}{natbib}
\PassOptionsToPackage{table}{xcolor}
\usepackage[preprint]{neurips_2026}

\usepackage[utf8]{inputenc}
\usepackage[T1]{fontenc}
\usepackage{hyperref}
\usepackage{url}
\usepackage{booktabs}
\usepackage{amsfonts}
\usepackage{amsmath}
\usepackage{amssymb}
\usepackage{nicefrac}
\usepackage{microtype}
\usepackage{xcolor}
\usepackage{graphicx}
\usepackage{multirow}
\usepackage{array}
\usepackage{algorithm}
\usepackage{algpseudocode}
\usepackage{enumitem}
\usepackage{xspace}
\usepackage{wrapfig}

\usepackage{caption}

\newcommand{\method}{CFGPatch\xspace}

\title{Exposing Vulnerabilities in Visible-Infrared VLMs: A Unified Geometric 
Adversarial Framework with Cross-Task Transferability}

\author{Xiang Chen \quad Yuxian Dong \quad Chao Li \quad Chengyin Hu \quad Jiaju Han\\
Fengyu Zhang \quad Yiwei Wei \quad Jiahuan Long \quad Jiujiang Guo}

\begin{document}

\maketitle

\begin{abstract}
Vision--language models (VLMs) have made substantial progress in multimodal understanding and achieve strong performance across diverse vision--language tasks. However, existing VLM security studies remain largely confined to single-modality settings, predominantly in the visible spectrum, leaving adversarial robustness in visible--infrared (VIS--IR) scenarios underexplored. This omission is consequential for real-world perception systems, where VIS--IR sensing is widely deployed to sustain reliable semantic understanding under challenging imaging conditions. Motivated by this underexplored cross-modal threat setting, we propose a curved-edge fractal geometric adversarial patch (CFGPatch), the first unified adversarial patch framework for cross-modal attacks against VIS--IR VLMs. Specifically, CFGPatch takes triangular fractal geometry as its base and transforms rigid straight-edged primitives into Bezier-curved elements, preserving fractal self-similarity for multi-scale adversarial effects while enhancing structural expressiveness through smoother contours, richer directional variation, and flexible shape deformation. Complementing this global geometric disruption, we introduce a modality-specific Fraser-spiral rendering mechanism that injects fine-grained texture distortions and misleading perceptual cues tailored to visible and infrared imagery. Through this geometry--appearance coupling, CFGPatch forms a dual-level perturbation: the curved fractal geometry disrupts global shape perception, while Fraser-spiral rendering corrupts local appearance interpretation via modality-specific texture interference. We further use expectation over transformation (EOT) to improve patch robustness against common image-level transformations. Extensive experiments show that CFGPatch effectively fools VIS--IR VLMs and outperforms standard patch baselines in both attack effectiveness and robustness. Moreover, adversarial samples optimized for the zero-shot classification task transfer well to image captioning and visual question answering (VQA), demonstrating the strong cross-task transferability and generalizability of CFGPatch across downstream tasks.
\end{abstract}

\section{Introduction}

VLMs have become a central paradigm for multimodal understanding, achieving strong performance across zero-shot classification, image captioning, and VQA \citep{radford2021clip,jia2021align,li2021albef,li2022blip,li2023blip2,alayrac2022flamingo,liu2023visual,chen2024internvl}. However, existing robustness studies on VLMs remain concentrated on single-modality settings, especially in the visible spectrum \citep{chen2023benchmarking,xie2025chain,shirnin2024analyzing,zhao2023vlmattack,zhang2025enhancing,ying2025jailbreak,wang2025vismodal}. This leaves VIS--IR robustness insufficiently understood, despite the broad use of dual-spectrum perception in real-world visual analysis, where visible imagery provides rich appearance cues while infrared sensing offers complementary information under poor illumination and degraded imaging conditions \citep{tang2022piafusion,zhang2020objectfusion}. Given the complementary evidence provided by the two spectra, perturbing only one branch yields an incomplete characterization of system vulnerability, motivating a stronger threat model in which a single perturbation family simultaneously disrupts semantic reasoning across both visible and infrared branches. Recent studies have explored cross-modal adversarial patches for visible–infrared scenarios, yet existing work remains detector-oriented, focusing on object detection or tracking rather than the more challenging semantic reasoning setting of VLMs \citep{wei2023unified,hu2025touap,long2025cdu}.

\begin{wrapfigure}{r}{0.46\textwidth}
    \vspace{-0.8em}
    \centering
    \includegraphics[width=0.44\textwidth]{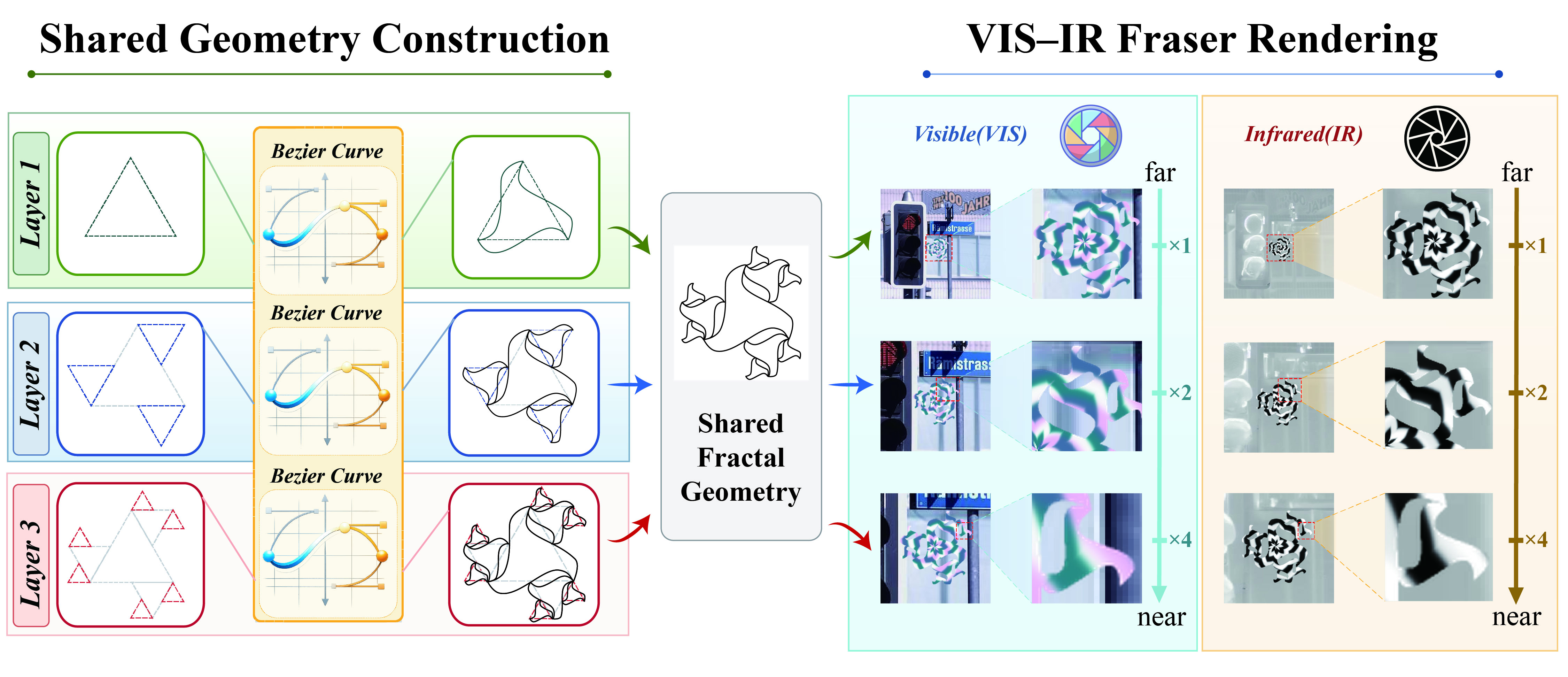}
    \caption{Shared curved-fractal construction and VIS--IR Fraser rendering of CFGPatch.}
    \label{fig:fractal_construction}
    \vspace{-0.8em}
\end{wrapfigure}

Moving from detector-oriented VIS--IR attacks to semantic attacks on VIS--IR VLMs introduces several intertwined challenges. A unified patch should preserve a shared geometric identity across spectra, since using unrelated shapes for visible and infrared inputs would collapse the setting into independent single-modality attacks. In parallel, this shared structure must remain sufficiently expressive to accommodate the distinct imaging characteristics of the two branches, which respond differently to the same spatial pattern. Beyond cross-spectral consistency, attacking VLMs also requires perturbations that affect semantic reasoning rather than only local visual evidence. This motivates a compact patch design capable of jointly modulating coarse shape cues and fine-grained appearance cues, consistent with prior studies suggesting that high-level reasoning relies on globally coherent visual evidence rather than only isolated pixels \citep{li2022finegrained,covert2025locality,zhang2025assessing}.

To address these challenges, we introduce \method, the first unified adversarial patch framework designed to expose cross-modal vulnerabilities in VIS--IR VLMs. Figure~\ref{fig:fractal_construction} illustrates the two core components of our design: a curved-edge triangular fractal geometry and modality-specific Fraser-spiral rendering. Specifically, we take triangular fractal geometry as the base structure and redesign its originally rigid straight-edged primitives into Bezier-curved elements, producing a compact multi-scale geometric carrier shared across spectra. The Fraser-spiral renderer then instantiates this shared carrier with visible- and infrared-specific appearance patterns. This geometry--appearance coupling yields a dual-level perturbation that disrupts global shape perception and local appearance interpretation within a unified cross-modal geometry.

We optimize CFGPatch in a compact parameter space using particle swarm optimization (PSO), and incorporate EOT to improve the robustness of the learned patch against common image-level transformations such as scale changes, rotations, blur, and appearance variations \citep{kennedy1995pso,athalye2018eot}. A paired attack is counted as successful only when the same shared-geometry patch fools both the visible and infrared branches simultaneously. We use zero-shot classification as the primary optimization task and directly reuse the resulting adversarial samples for image captioning and VQA without task-specific re-optimization.

Overall, our work extends VIS--IR adversarial patch research from detector-oriented evasion to a broader semantic threat model for VLMs. Our contributions are summarized as follows:
\begin{itemize}
    \item To the best of our knowledge, we present \method, the first unified adversarial patch framework for VIS--IR VLMs. The framework is instantiated by transforming triangular fractal geometry into a shared curved-edge geometric carrier and coupling it with Fraser-spiral-based modality-specific rendering, thereby integrating cross-spectral structural consistency with spectrum-adaptive appearance interference.
    
    \item Through extensive experiments on zero-shot classification, image captioning, and VQA, we show that \method consistently surpasses representative VIS--IR patch baselines and achieves strong simultaneous attack performance across visible and infrared branches.
    
    \item Comprehensive ablation studies and detailed analysis of CFGPatch further reveal the contributions of key design choices, including curved-edge fractal geometry, modality-specific Fraser rendering, fractal depth, and optimization budget, thereby systematically validating the robustness and cross-task transferability of the proposed framework.
\end{itemize}

\begin{figure*}[t]
    \centering
    \includegraphics[width=\textwidth]{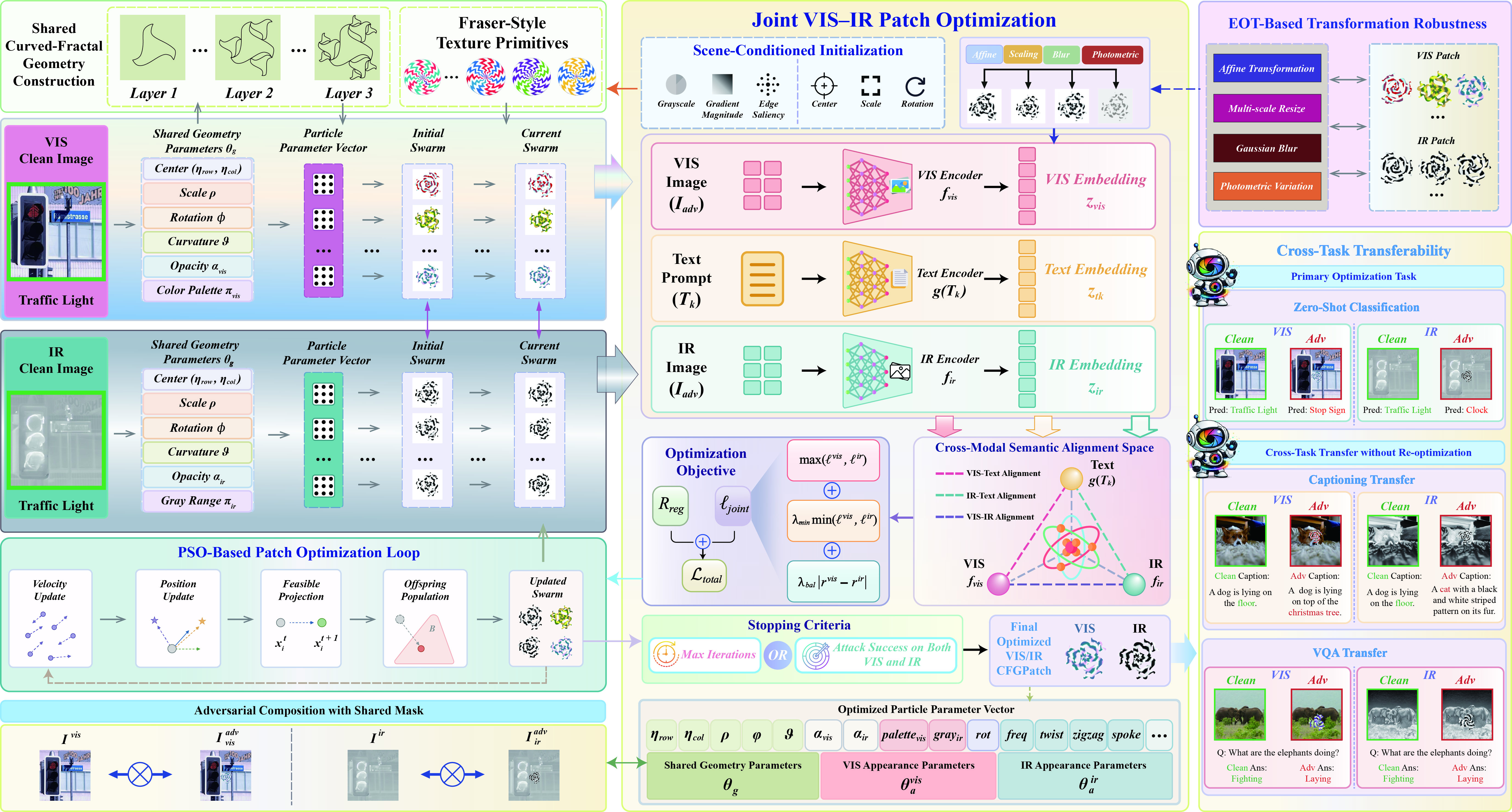}
    \caption{Overview of \method: a shared curved--edge fractal patch for VIS--IR inputs is optimized under PSO with EOT and transferred across tasks.}
    \label{fig:framework}
\end{figure*}

\section{Related Work}

\paragraph{Robustness of vision--language models.}
A growing body of work has shown that VLMs are vulnerable to both textual and visual perturbations. Prompt design can substantially affect zero-shot behavior \citep{zhou2022coop,zhou2022cocoop,khattak2023maple,khattak2023self}, while image-space adversarial perturbations can distort multimodal representation geometry and degrade downstream performance \citep{chen2023benchmarking,xie2025chain,shirnin2024analyzing,zhao2023vlmattack,zhang2025enhancing,ying2025jailbreak}. Recent corruption and benchmark studies further indicate that robustness under realistic visual shifts remains limited even for strong large-scale VLMs \citep{wang2025vismodal,qiu2025benchmarking,usama2025analysing,ying2024mmt}. These findings suggest that the robustness of VLMs depends not only on visual recognition accuracy, but also on the stability of cross-modal alignment. However, most existing studies focus on single-modality settings, especially the visible spectrum, leaving robustness in visible--infrared scenarios largely unexplored.

\paragraph{Geometry priors and structured perturbations.}
Compared with unconstrained pixel-space perturbations, structured perturbations impose compact and interpretable priors on the attack space. Adversarial patches demonstrate that localized structured patterns can reliably influence model predictions \citep{brown2017patch,eykholt2018rp2,chen2018shapeshifter,liu2018dpatch}, while transformation-aware optimization such as EOT improves robustness under realistic variation \citep{athalye2018eot}. More broadly, geometric priors are appealing because they provide expressive yet controllable structure for compact perturbations. In particular, recursive self-similarity in fractal geometry offers a natural way to generate multi-scale patterns from a small parameter set \citep{lutz2008dimensions}, making it well suited to attacks that aim to affect both coarse visual organization and fine local cues. This perspective is also consistent with recent physically grounded perturbation studies based on natural shadows, light, illumination transformation, and 3D environmental variation \citep{zhong2022shadow,hsiao2024natural,liu2025ita,ruan2025advdreamer}.

\paragraph{Cross-modal visible--infrared adversarial attacks.}
Recent studies \citep{wei2023unified,hu2025touap,long2025cdu} have primarily focused on detector-centric visible--infrared adversarial attacks, where patch methods optimize shared shape cues or modality-adapted appearance to compromise object detectors under varying imaging conditions. In parallel, recent progress in infrared vision--language modeling has started to extend multimodal semantic understanding from visible imagery to the infrared domain \citep{jiang2024infraredllava,cao2025irgpt}, and very recent work has further exposed the vulnerability of infrared VLMs through universal adversarial patches \citep{hu2026ucgp}. In contrast, our work studies a stronger setting in which a single structured patch preserves one shared geometry while simultaneously attacking both visible and infrared VLMs.

\section{Method}

\subsection{Overview}

Given paired visible and infrared images $(I^{\mathrm{vis}}, I^{\mathrm{ir}})$ with ground-truth label $y$, we aim to construct a single cross-modal patch family that enforces shared geometric support across spectra while allowing modality-specific appearance rendering. Figure~\ref{fig:framework} summarizes \method{}, which combines shared geometric disruption with modality-specific appearance interference: a common patch mask provides spectrum-consistent spatial structure, while visible Fraser-style and infrared grayscale textures introduce spectrum-dependent perturbations. The patch is optimized with a joint VIS--IR objective for zero-shot classification, and the resulting adversarial samples are subsequently evaluated on image captioning and VQA.

\subsection{Problem Definition}

We formulate the attack under the zero-shot VIS--IR classification task, where each paired image sample is associated with a ground-truth label $y\in\mathcal{Y}=\{1,\dots,K\}$. For each modality $m\in\{\mathrm{vis},\mathrm{ir}\}$, the VLM performs recognition by comparing the image representation with class-level text prompts. The image--text similarity score for class $k$ is computed as:
\begin{equation}
s_k^{(m)}(I)
=
\left\langle f_m(I),\, g(T_k)\right\rangle
\label{eq:similarity}
\end{equation}
where $f_m(\cdot)$ denotes the image encoder for modality $m$, $g(\cdot)$ is the shared text encoder, and $T_k$ is the text prompt associated with class $k$. The predicted label for modality $m$ is then given by:
\begin{equation}
\hat{y}^{(m)}(I)
=
\arg\max_{k\in\mathcal{Y}} s_k^{(m)}(I)
\label{eq:prediction}
\end{equation}
Here, $s_k^{(m)}(I)$ measures the alignment between image $I$ and class prompt $T_k$, while $\hat{y}^{(m)}(I)$ denotes the resulting zero-shot prediction.

We evaluate only paired samples for which both clean branches are correctly classified, so that attack success is not inflated by pre-existing recognition failures:
\begin{equation}
\hat{y}^{(\mathrm{vis})}\!\left(I^{\mathrm{vis}}\right)=y,
\qquad
\hat{y}^{(\mathrm{ir})}\!\left(I^{\mathrm{ir}}\right)=y
\label{eq:clean-filter}
\end{equation}
After applying the adversarial patch, a paired attack is counted as successful only when both the visible and infrared branches are misclassified:
\begin{equation}
\hat{y}^{(\mathrm{vis})}\!\left(I_{\mathrm{adv}}^{\mathrm{vis}}\right)\neq y,
\qquad
\hat{y}^{(\mathrm{ir})}\!\left(I_{\mathrm{adv}}^{\mathrm{ir}}\right)\neq y
\label{eq:success}
\end{equation}
This conjunction-based criterion defines a strict VIS--IR attack protocol, where fooling only one spectrum is insufficient for cross-modal success.

\subsection{Shared Curved-Triangle Fractal Geometry}

The patch support is defined by a recursively constructed curved-triangle fractal layout. Starting from a canonical equilateral triangle, we generate a hierarchy of guide triangles and retain the level-$1$, level-$2$, and level-$3$ structures jointly, resulting in a compact yet multi-scale support region. This recursive design is important because the self-similarity of fractal geometry introduces repeated patterns across scales, enabling the patch to simultaneously interfere with coarse object-level structure and fine-grained local evidence.

To further enhance geometric expressiveness, each triangle edge is parameterized by a cubic Bezier curve instead of a straight segment. For an edge with endpoints $p_0,p_3\in\mathbb{R}^2$ and control points $p_1,p_2\in\mathbb{R}^2$, the curve is defined as:
\begin{equation}
B(t)
=
(1-t)^3 p_0
+
3(1-t)^2 t\, p_1
+
3(1-t)t^2 p_2
+
t^3 p_3
\label{eq:bezier}
\end{equation}
where $t\in[0,1]$ is the curve parameter, and $p_1$ and $p_2$ determine the local tangent and curvature. Compared with rigid polygonal edges, this parameterization produces smoother boundaries, richer directional variation, and more expressive shape deformations.

We denote the shared geometry parameters by:
\begin{equation}
\theta_g
=
\{\eta_{\mathrm{row}},\eta_{\mathrm{col}},\rho,\phi,\vartheta,\theta_{\mathrm{shape}}\}
\label{eq:theta-g}
\end{equation}
where $\eta_{\mathrm{row}}$ and $\eta_{\mathrm{col}}$ denote the patch center coordinates along the image row and column directions, respectively, $\rho$ is the global scale, $\phi$ is the in-plane rotation, $\vartheta$ controls the overall curvature strength, and $\theta_{\mathrm{shape}}$ collects the remaining shape variables induced by the recursive construction. By rasterizing and unioning all retained curved triangles, we obtain a shared geometry mask:
\begin{equation}
M_g(\theta_g)\in[0,1]^{H\times W}
\label{eq:mask}
\end{equation}
where $H$ and $W$ denote the image height and width. The same mask $M_g(\theta_g)$ is applied to both visible and infrared modalities, thereby directly enforcing the shared-geometry constraint.

\begin{table*}[t]
    \centering
    \scriptsize
    \setlength{\tabcolsep}{2.8pt}
    \renewcommand{\arraystretch}{1.08}
    \caption{Attack results (\%) for zero-shot classification on paired VIS--IR inputs. Higher values indicate stronger attack success.}
    \resizebox{\textwidth}{!} {
    \begin{tabular}{lcccccccccccc}
        \toprule
        \multirow{2}{*}{Method}
        & \multicolumn{3}{c}{OpenAI CLIP ViT-L/14}
        & \multicolumn{3}{c}{OpenCLIP ViT-B/16}
        & \multicolumn{3}{c}{Meta-CLIP ViT-L/14}
        & \multicolumn{3}{c}{EVA-CLIP ViT-G/14} \\
        \cmidrule(lr){2-4} \cmidrule(lr){5-7} \cmidrule(lr){8-10} \cmidrule(lr){11-13}
        & VIS & IR & VIS--IR
        & VIS & IR & VIS--IR
        & VIS & IR & VIS--IR
        & VIS & IR & VIS--IR \\
        \midrule
        Unified Patch & 21.15 & 36.54 & 14.10 & 24.05 & 40.51 & 17.72 & 10.22 & 13.44 &  8.06 & 16.26 & 21.18 & 12.81 \\
        TOUAP         & 38.46 & 46.79 & 25.64 & 43.04 & 69.62 & 36.10 & 29.03 & 55.37 & 25.27 & 39.41 & 40.87 & 27.09 \\
        Ours (Random) & 61.53 & 65.38 & 54.49 & 72.15 & 88.61 & 65.19 & 63.44 & 66.13 & 59.14 & 65.02 & 73.40 & 64.53 \\
        \rowcolor{gray!15}
        \textbf{Ours} & \textbf{69.54} & \textbf{84.48} & \textbf{67.82}
                      & \textbf{80.38} & \textbf{92.41} & \textbf{77.22}
                      & \textbf{72.58} & \textbf{80.65} & \textbf{68.28}
                      & \textbf{77.34} & \textbf{79.80} & \textbf{73.89} \\
        \bottomrule
    \end{tabular}}
\label{tab:cls_main}
\end{table*}

\subsection{Modality-Specific Fraser Rendering}

While the patch geometry is shared across modalities, its appearance is rendered in a modality-specific manner. This design is motivated by a simple physical intuition: a single patch carrier can preserve a common geometric identity while exhibiting different spectral responses under visible and infrared imaging. Based on this observation, we build a Fraser-spiral-based rendering module on top of the shared geometry mask.

Let $\theta_a^{\mathrm{vis}}$ and $\theta_a^{\mathrm{ir}}$ denote the appearance parameters for the visible and infrared branches, respectively. Both branches are driven by a common Fraser-style coordinate field, whose structure is controlled by factors such as rotation, phase, ring frequency, twist, zigzag amplitude, spoke number, and sharpness. On this basis, the visible branch maps the field to a color palette, whereas the infrared branch maps it to a grayscale intensity range. The resulting patch textures are defined as:
\begin{align}
P^{\mathrm{vis}}
&=
R_{\mathrm{vis}}\!\left(\theta_g,\theta_a^{\mathrm{vis}}\right)
\label{eq:render-vis} \\
P^{\mathrm{ir}}
&=
R_{\mathrm{ir}}\!\left(\theta_g,\theta_a^{\mathrm{ir}}\right)
\label{eq:render-ir}
\end{align}
where $R_{\mathrm{vis}}(\cdot)$ and $R_{\mathrm{ir}}(\cdot)$ denote the rendering functions for the visible and infrared modalities, respectively. In both branches, the spatial support is inherited from the shared geometry parameters $\theta_g$, while the texture appearance is controlled by modality-specific parameters. The adversarial visible and infrared images are then obtained by alpha blending:
\par\vspace{-1em}

\begingroup
\setlength{\abovedisplayskip}{3pt}
\setlength{\belowdisplayskip}{3pt}
\setlength{\abovedisplayshortskip}{3pt}
\setlength{\belowdisplayshortskip}{3pt}
\begin{gather}
I_{\mathrm{adv}}^{\mathrm{vis}}
=
\left(1-\alpha_{\mathrm{vis}} M_g\right)\odot I^{\mathrm{vis}}
+
\alpha_{\mathrm{vis}} M_g \odot P^{\mathrm{vis}}
\label{eq:adv-vis}
\\[0.5em]
I_{\mathrm{adv}}^{\mathrm{ir}}
=
\left(1-\alpha_{\mathrm{ir}} M_g\right)\odot I^{\mathrm{ir}}
+
\alpha_{\mathrm{ir}} M_g \odot P^{\mathrm{ir}}
\label{eq:adv-ir}
\end{gather}
\endgroup

\vspace{-0.35em}
\noindent where $\odot$ denotes element-wise multiplication, $M_g=M_g(\theta_g)$ is the shared geometry mask, and $\alpha_{\mathrm{vis}},\alpha_{\mathrm{ir}}\in[0,1]$ are modality-specific opacity coefficients. This construction yields a two-level perturbation mechanism: the outer curved fractal geometry disrupts global shape perception, while the inner Fraser-style texture interferes with local appearance cues.

\subsection{Joint Cross-Modal Objective}

Our goal is to drive the visible and infrared branches toward simultaneous failure, rather than allowing the optimization to overfit only one modality. For each modality $m\in\{\mathrm{vis},\mathrm{ir}\}$, let $s_y^{(m)}$ denote the similarity score of the ground-truth class $y$, and let the strongest competing score be defined as:
\begin{equation}
s_{\max\setminus y}^{(m)}
=
\max_{k\neq y} s_k^{(m)}
\label{eq:max-comp}
\end{equation}
Based on this quantity, we define a margin-based loss for each modality as:
\begin{equation}
\ell^{(m)}
=
\mathrm{softplus}\!\Bigl(
s_y^{(m)} - s_{\max\setminus y}^{(m)} + \delta
\Bigr)
\label{eq:modality-loss}
\end{equation}
where $\delta>0$ is a safety margin. Intuitively, $\ell^{(m)}$ becomes small when the strongest incorrect class already exceeds the ground-truth class by a sufficient margin. To more directly characterize attack progress, we further define the attack reward as:
\begin{equation}
r^{(m)}
=
s_{\max\setminus y}^{(m)} - s_y^{(m)}
\label{eq:reward}
\end{equation}
which measures the extent to which the most competitive incorrect class dominates the true class.

To couple the two modalities, we construct the following joint objective:
\begin{equation}
\ell_{\mathrm{joint}}
=
\max\!\bigl(\ell^{(\mathrm{vis})},\,\ell^{(\mathrm{ir})}\bigr)
+
\lambda_{\min}\,
\min\!\bigl(\ell^{(\mathrm{vis})},\,\ell^{(\mathrm{ir})}\bigr)
+
\lambda_{\mathrm{bal}}
\bigl|r^{(\mathrm{vis})}-r^{(\mathrm{ir})}\bigr|
\label{eq:joint-loss}
\end{equation}
where $\lambda_{\min}>0$ controls the contribution of the easier modality, and $\lambda_{\mathrm{bal}}>0$ penalizes imbalanced attack progress across spectra. In our implementation, we set $\lambda_{\min}=0.55$. The first term emphasizes the harder branch, the second term preserves optimization pressure on the easier branch, and the balance term discourages one-sided fooling.

To improve robustness under realistic deployment variations, we optimize the patch under expectation over transformation. Let $\tau\sim\mathcal{T}$ denote a random transformation sampled from a distribution $\mathcal{T}$ over affine perturbations, scale changes, blur, and photometric variations. The final optimization problem is formulated as:
\begin{equation}
\min_{\theta_g,\theta_a^{\mathrm{vis}},\theta_a^{\mathrm{ir}}}
\;
\mathbb{E}_{\tau\sim\mathcal{T}}
\left[
\ell_{\mathrm{joint}}
\Bigl(
\tau(I_{\mathrm{adv}}^{\mathrm{vis}}),
\tau(I_{\mathrm{adv}}^{\mathrm{ir}})
\Bigr)
\right]
+
\lambda_{\mathrm{reg}}\,\mathcal{R}_{\mathrm{reg}}
\label{eq:full-objective}
\end{equation}
where $\mathcal{R}_{\mathrm{reg}}$ denotes the overall regularization term and $\lambda_{\mathrm{reg}}>0$ controls its strength. Specifically, $\mathcal{R}_{\mathrm{reg}}$ consists of lightweight penalties on patch area, mask smoothness, geometry stabilization, and texture smoothness, which help constrain the patch while preserving sufficient attack flexibility. We use the notation $\mathcal{R}_{\mathrm{reg}}$ instead of $\mathcal{R}$ to avoid ambiguity with the rendering operators $R_{\mathrm{vis}}$ and $R_{\mathrm{ir}}$.

\begin{figure*}[t]
    \centering
    \includegraphics[width=\textwidth]{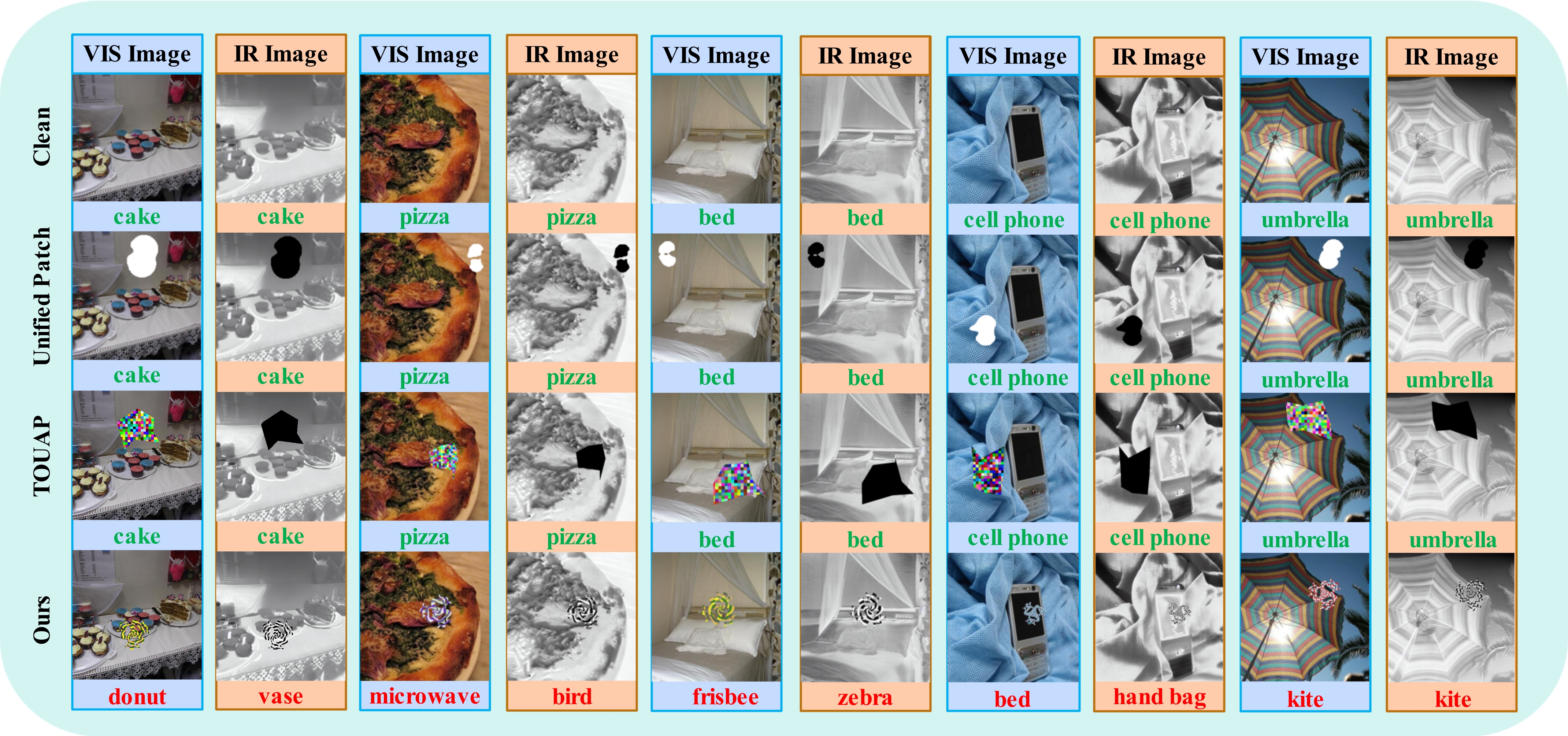}
    \caption{Qualitative classification examples in the VIS--IR setting. From top to bottom: Clean, Unified Patch, TOUAP, and Ours (\method); left: VIS Image, right: IR Image. Green and red text denote correct and adversarial predictions, respectively.}
    \label{fig:qualitative_cls}
\end{figure*}

\subsection{Scene-Conditioned Initialization and PSO}

Since our patch involves recursive rasterized geometry, discrete palette choices, and non-smooth Fraser rendering, gradient-based optimization is unreliable in this parameter space. We therefore optimize the shared geometry and modality-specific appearance parameters with PSO.

To provide a stable and informative starting point, we use a scene-conditioned initialization strategy. The visible and infrared images are first converted to grayscale, their gradient magnitudes are averaged to capture cross-modal edge saliency, and the initial patch center is estimated from regions with concentrated structural responses. A lightweight coarse search over candidate centers, scales, and rotations further refines this state, biasing the patch toward structurally informative regions that are consistently salient across both modalities.

Each particle encodes the full parameter vector, including shared geometry and modality-specific appearance parameters. For particle $i$ at iteration $t$, let $x_i^{\,t}$ and $v_i^{\,t}$ denote its position and velocity, $b_i^{\,t}$ its personal best, and $b_\star^{\,t}$ the global best. The PSO update is defined as:
\begin{gather}
v_i^{\,t+1}
=
\omega\, v_i^{\,t}
+
c_1 \xi_{1,i}^{\,t}\odot\bigl(b_i^{\,t}-x_i^{\,t}\bigr)
+
c_2 \xi_{2,i}^{\,t}\odot\bigl(b_\star^{\,t}-x_i^{\,t}\bigr)
\label{eq:pso-v}
\\
x_i^{\,t+1}
=
\Pi_{\mathcal{B}}\!\left(x_i^{\,t}+v_i^{\,t+1}\right)
\label{eq:pso-x}
\end{gather}
where $\omega$ is the inertia weight, $c_1$ and $c_2$ are the cognitive and social coefficients, $\xi_{1,i}^{\,t}$ and $\xi_{2,i}^{\,t}$ are element-wise random vectors sampled from $[0,1]$, and $\Pi_{\mathcal{B}}$ projects parameters onto the feasible domain $\mathcal{B}$. At each iteration, particles are rendered under the shared-geometry constraint, evaluated with sampled EOT transformations, and updated according to Eqs.~\eqref{eq:pso-v}--\eqref{eq:pso-x}. The search stops early once both modalities are successfully fooled. Pseudo-code is provided in the supplementary material.

\begin{table*}[t]
    \centering
    \scriptsize
    \setlength{\tabcolsep}{3.8pt}
    \renewcommand{\arraystretch}{1.12}
    \caption{Attack results (\%) on image captioning for paired VIS--IR inputs. Values in parentheses indicate the performance drop relative to the corresponding clean baseline.}
    \label{tab:cap_main}
    \resizebox{\textwidth}{!}{
    \begin{tabular}{cccccccccc}  
        \toprule
        \multirow{2}{*}{Image Encoder} & \multirow{2}{*}{Models}
        & \multicolumn{2}{c}{Clean}
        & \multicolumn{2}{c}{Unified Patch}
        & \multicolumn{2}{c}{TOUAP}
        & \multicolumn{2}{c}{Ours} \\
        \cmidrule(lr){3-4} \cmidrule(lr){5-6} \cmidrule(lr){7-8} \cmidrule(lr){9-10}
        &  & VIS & IR
           & VIS$\downarrow$ & IR$\downarrow$
           & VIS$\downarrow$ & IR$\downarrow$
           & VIS$\downarrow$ & IR$\downarrow$ \\
        \midrule
        \multirow{4}{*}{\shortstack[c]{OpenAI CLIP\\ViT-L/14}}
        & LLaVA-1.5 (7B)             & 78.53 & 65.40 & 76.00 (2.53) & 58.87 (6.53) & 71.95 (6.58) & 55.15 (10.25) & \textbf{65.39 (13.14)} & \textbf{50.25 (15.15)} \\
        & LLaVA-1.6 (7B)            & 82.37 & 68.00 & 81.12 (1.25) & 57.17 (10.83) & 71.25 (11.12) & 56.45 (11.55) & \textbf{67.20 (15.17)} & \textbf{51.54 (16.46)} \\
        & OpenFlamingo (3B)          & 79.46 & 67.36 & 77.94 (1.52) & 61.60 (5.76) & 71.20 (8.26) & 56.94 (10.42) & \textbf{64.53 (14.93)} & \textbf{47.31 (20.05)} \\
        & Blip-2 FlanT5XL ViT-L (3.4B) & 85.68 & 62.19 & 83.95 (1.73) & 57.09 (5.10) & 79.40 (6.28) & 54.26 (7.93) & \textbf{75.65 (10.03)} & \textbf{49.12 (13.07)} \\
        \midrule
        \multirow{2}{*}{\shortstack[c]{EVA-CLIP\\ViT-G/14}}
        & Blip-2 FlanT5XL (4.1B)       & 75.38 & 65.03 & 71.89 (3.49) & 57.38 (7.65) & 63.84 (11.54) & 56.83 (8.20) & \textbf{60.92 (14.46)} & \textbf{48.97 (16.06)} \\
        & InstructBLIP FlanT5XL (4.1B) & 77.56 & 68.44 & 75.64 (1.92) & 65.73 (2.71) & 72.11 (5.45) & 59.37 (9.07) & \textbf{69.81 (7.75)} & \textbf{52.91 (15.53)} \\
        \bottomrule
    \end{tabular}}
\end{table*}

\section{Experiments}
\label{sec:experiments}

\subsection{Experimental Settings}
\label{sec:exp_setup}

\paragraph{Datasets.}
We evaluate \method{} on a paired VIS--IR benchmark built from 300 MS COCO images spanning 30 semantic categories \citep{lin2014coco,chen2015cococaptions}. These categories define the label space for zero-shot classification through standard prompt templates \citep{zhou2022coop,khattak2023self}. Following the approach proposed in \citep{han2024dravit}, we convert each visible image into a semantically aligned infrared counterpart, preserving scene-level semantics while mapping the visible appearance to the infrared domain. For image captioning and VQA, we use the adversarial samples generated from the zero-shot classification attack as task inputs, together with the corresponding captions and question-answer annotations for evaluation. The prompt templates used for captioning and VQA are provided in the supplementary material.

\paragraph{Models and Infrared Adaptation Protocol.}
For zero-shot classification, we evaluate OpenCLIP ViT-B/16 \citep{cherti2023openclip}, Meta-CLIP ViT-L/14 \citep{xu2023metaclip}, EVA-CLIP ViT-G/14 \citep{fang2023eva}, and OpenAI CLIP ViT-L/14 \citep{radford2021clip}. For captioning and VQA, we evaluate LLaVA-1.5  \citep{liu2023visual}, LLaVA-1.6  \citep{liu2024llava}, OpenFlamingo  \citep{awadalla2023openflamingo}, BLIP-2 \citep{li2023blip2}, and InstructBLIP \citep{dai2023instructblip}. For the infrared branch, all backbones are adapted from their visible counterparts under a unified IR-VLM protocol. We use an infrared semantic corpus \citep{jiang2024infraredllava} and perform lightweight LoRA adaptation \citep{hu2022lora} from natural-image-pretrained weights. Generative models are fine-tuned with next-token prediction, while contrastive models use InfoNCE \citep{oord2018representation} , preserving architectural comparability across modalities.

\paragraph{Evaluation Metrics.}  
For zero-shot classification, we report the attack success rate (ASR) for the VIS and IR branches individually, and for both modalities simultaneously, which represents the success of the dual-modality attack. For image captioning and VQA, we follow the LLM-as-Judge protocol \citep{bavaresco2025llms}, using GPT-5 to evaluate caption semantic consistency and answer correctness, reporting the attack results together with its drop relative to the clean baseline.

\paragraph{Baselines.}
For zero-shot classification, we compare \method{}, along with its variants optimized via random search (Radom) and PSO, to two representative VIS--IR adversarial patch baselines: Unified Patch \citep{wei2023unified} and TOUAP \citep{hu2025touap}, highlighting the effectiveness of our approach in generating adversarial patches. We evaluate all baselines under the same protocol for additional tasks, including image captioning and VQA.

\begin{figure*}[t]
    \centering
    \includegraphics[width=\textwidth]{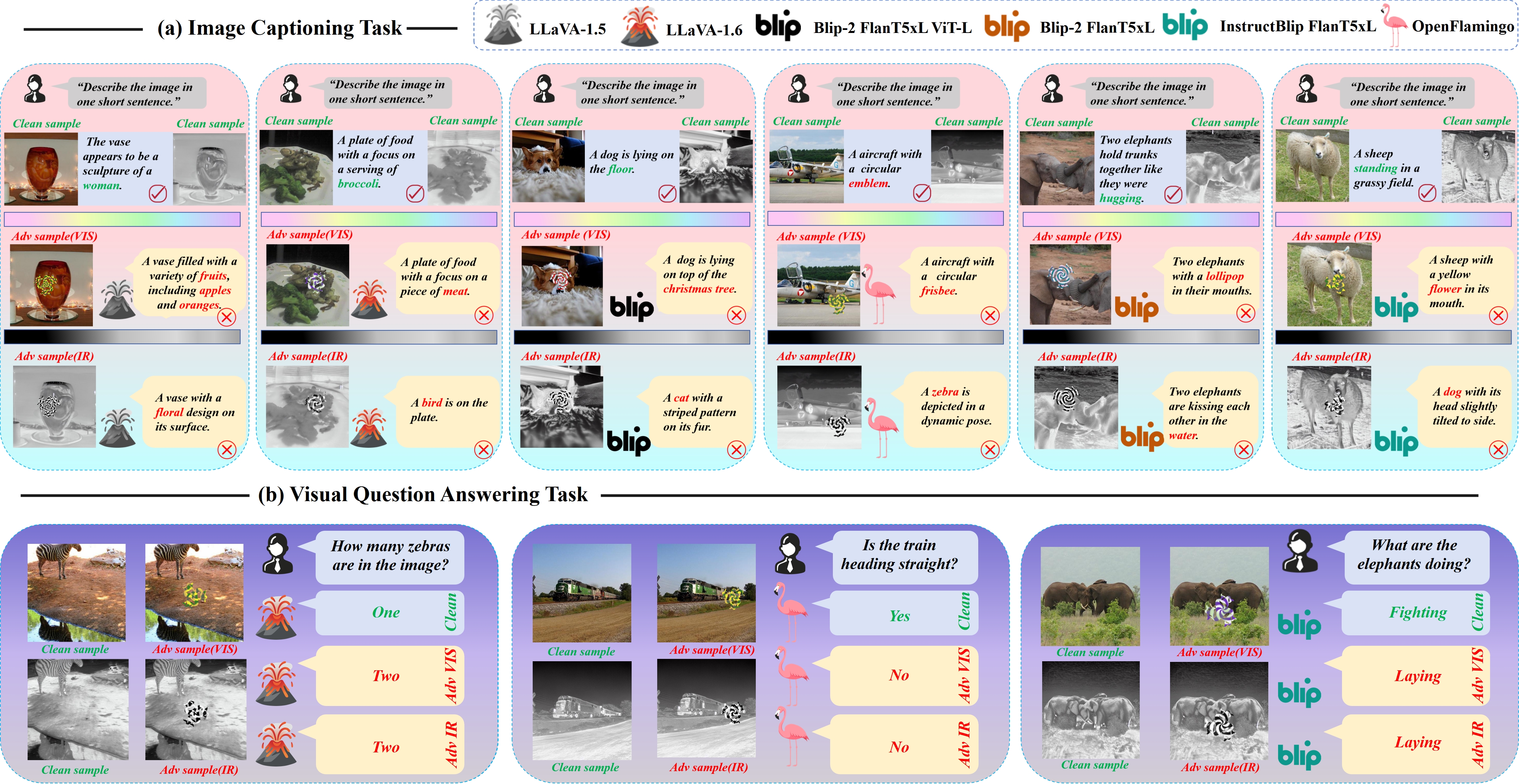}
   \caption{Qualitative examples. (a) Captioning under clean and adversarial VIS--IR inputs. (b) VQA under clean and adversarial conditions for both visible and infrared branches.}
    \label{fig:qualitative_cap_vqa}
\end{figure*}

\paragraph{Implementation Details.}
All attacks are optimized using particle swarm optimization (PSO) under EOT \citep{kennedy1995pso,athalye2018eot}. In our experiments, we set the swarm size to 20, the optimization budget to 150 iterations, and the number of EOT samples to 2. The patch is parameterized as a depth-3 curved fractal structure, with optimization variables covering shared geometry, opacity, visible palette, infrared grayscale range, and Fraser-texture parameters.

\begin{table*}[t]
    \centering
    \scriptsize
    \setlength{\tabcolsep}{3.2pt}
    \renewcommand{\arraystretch}{1.10}
    \caption{Attack results (\%) on visual question answering (VQA) for paired VIS--IR inputs. Values in parentheses indicate the performance drop relative to the corresponding clean baseline.}
    \label{tab:vqa_main}
    \resizebox{\textwidth}{!}{
    \begin{tabular}{cccccccccc}  
        \toprule
        \multirow{2}{*}{Image Encoder} & \multirow{2}{*}{Models}
        & \multicolumn{2}{c}{Clean}
        & \multicolumn{2}{c}{Unified Patch}
        & \multicolumn{2}{c}{TOUAP}
        & \multicolumn{2}{c}{Ours} \\
        \cmidrule(lr){3-4} \cmidrule(lr){5-6} \cmidrule(lr){7-8} \cmidrule(lr){9-10}
        &  & VIS & IR
           & VIS$\downarrow$ & IR$\downarrow$
           & VIS$\downarrow$ & IR$\downarrow$
           & VIS$\downarrow$ & IR$\downarrow$ \\
        \midrule
        \multirow{4}{*}{\shortstack[c]{OpenAI CLIP\\ViT-L/14}}
        & LLaVA-1.5 (7B)             & 65.38 & 62.27 & 61.97 (3.41) & 58.73 (3.54) & 59.24 (6.14) & 54.74 (7.53) & \textbf{52.81 (12.57)} & \textbf{49.80 (12.47)} \\
        & LLaVA-1.6 (7B)            & 61.75 & 54.64 & 59.11 (2.64) & 53.01 (1.63) & 56.24 (5.51) & 49.12 (5.52) & \textbf{50.12 (11.63)} & \textbf{42.58 (12.06)} \\
        & OpenFlamingo (3B)          & 68.26 & 63.58 & 66.74 (1.52) & 60.73 (2.85) & 64.11 (4.15) & 57.46 (6.12) & \textbf{55.91 (12.35)} & \textbf{53.02 (10.56)} \\
        & Blip-2 FlanT5XL ViT-L (3.4B) & 57.07 & 52.46 & 53.92 (3.15) & 48.70 (3.76) & 50.65 (6.42) & 45.72 (6.74) & \textbf{42.77 (14.30)} & \textbf{39.26 (13.20)} \\
        \midrule
        \multirow{2}{*}{\shortstack[c]{EVA-CLIP\\ViT-G/14}}
        & Blip-2 FlanT5XL (4.1B)       & 67.98 & 63.15 & 65.68 (2.30) & 59.40 (3.75) & 62.09 (5.89) & 54.58 (8.57) & \textbf{51.97 (16.01)} & \textbf{45.90 (17.25)} \\
        & InstructBLIP FlanT5XL (4.1B) & 68.41 & 64.72 & 66.73 (1.68) & 61.45 (3.27) & 62.78 (5.63) & 56.87 (7.85) & \textbf{56.90 (11.51)} & \textbf{51.60 (13.12)} \\
        \bottomrule
    \end{tabular}}
\end{table*}

\subsection{Zero-shot Classification}
\label{sec:cls}

Zero-shot classification serves as the primary evaluation task, providing a direct measure of whether a single shared patch can jointly fool both visible and infrared recognition. Table~\ref{tab:cls_main} reports the results on paired VIS--IR inputs, where the VIS and IR columns show the attack success rate for the visible and infrared branches individually, and VIS--IR indicates the success rate when both modalities are fooled simultaneously. As shown, \method{} achieves the highest ASR on both VIS and IR, as well as the highest joint success in VIS--IR, consistently outperforming Unified Patch and TOUAP across all four CLIP backbones. Qualitative examples are shown in Figure~\ref{fig:qualitative_cls}.

\subsection{Image Captioning Evaluation}
\label{sec:cap}

The adversarial VIS--IR pairs optimized on zero-shot classification are further applied to open-ended image captioning to assess cross-task transfer. As reported in Table~\ref{tab:cap_main}, \method{} causes the largest semantic degradation on both visible and infrared inputs across all evaluated captioning models. This result indicates that the attack effect is not limited to category-level decisions, but extends to free-form language generation by disrupting the visual semantics required for coherent caption construction. Representative qualitative examples for image captioning are shown in Figure~\ref{fig:qualitative_cap_vqa} (a).

\subsection{Visual Question Answering Evaluation}
\label{sec:vqa}

We also examine visual question answering, where correct responses require image-conditioned semantic reasoning rather than only visual recognition. Using the same adversarial VIS--IR pairs, Table~\ref{tab:vqa_main} shows that \method{} again produces the strongest performance drop across modalities and models. Compared with Unified Patch and TOUAP, the larger degradation suggests that the shared-geometry patch interferes with semantic grounding and reasoning beyond closed-set classification. Together with the captioning results, these findings show that \method{} induces transferable cross-task semantic disruption. Representative examples are shown in Figure~\ref{fig:qualitative_cap_vqa} (b).

\subsection{Ablation and Analysis}
\label{sec:ablation}

\paragraph{Component-level Ablation.}
We evaluate the contributions of Fraser rendering and Bezier boundary deformation. Removing either consistently reduces joint VIS--IR attack success, confirming the importance of both modality-specific texture and curved-edge geometry for effective attacks.

\begin{wrapfigure}[18]{r}{0.45\textwidth}
    \vspace{-0.6em}
    \centering
    \includegraphics[width=0.36\textwidth]{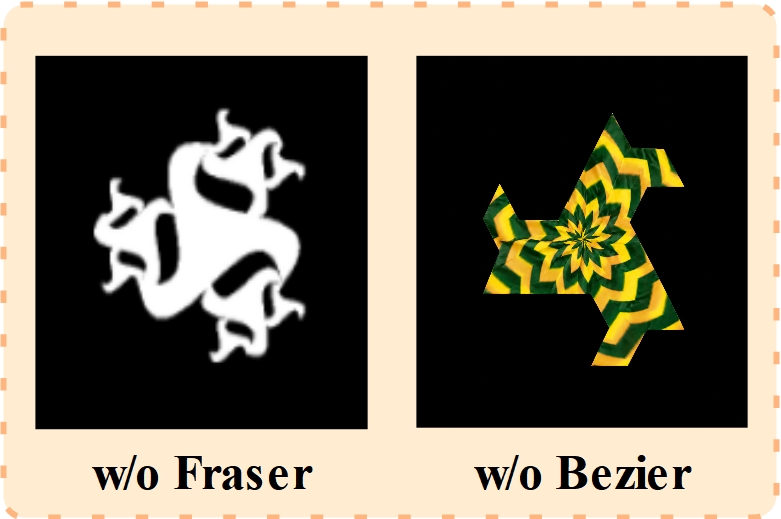}
    \caption{Patch variants in component ablation.}
    \label{fig:component_patch_visual}
    \vspace{0.2em}

    \captionof{table}{Component-level ablation results.}
    \label{tab:component_ablation}
    \vspace{0.2em}
    \scriptsize
    \setlength{\tabcolsep}{2.0pt}
    \renewcommand{\arraystretch}{1.02}
    \resizebox{0.43\textwidth}{!}{
    \begin{tabular}{lcccc}
        \toprule
        Methods & OpenAI CLIP & OpenCLIP & Meta-CLIP & EVA-CLIP \\
        & ViT-L/14 & ViT-B/16 & ViT-L/14 & ViT-G/14 \\
        \midrule
        w/o Fraser & 56.82 & 55.62 & 50.54 & 64.38 \\
        w/o Bezier & 57.19 & 60.45 & 62.73 & 67.12 \\
        CFGPatch & \textbf{67.82} & \textbf{77.22} & \textbf{68.28} & \textbf{73.89} \\
        \bottomrule
    \end{tabular}}
    \vspace{-1.0em}
\end{wrapfigure}

As summarized in Table~\ref{tab:component_ablation}, both ablated variants lead to clear performance degradation, indicating that CFGPatch relies on the coupling between spectrum-adaptive appearance rendering and expressive geometric support. Figure~\ref{fig:component_patch_visual} further illustrates the corresponding patch variants. Removing Fraser rendering weakens texture-level interference, while replacing Bezier curves reduces boundary expressiveness and limits the geometric flexibility of the patch.

\paragraph{Fractal Depth Ablation.}
We next analyze the effect of recursive fractal depth.
As shown in Figure~\ref{fig:depth_ablation}, deeper fractal recursion generally improves attack effectiveness, supporting the benefit of multi-scale geometric disruption for perturbing semantic representations across visible and infrared modalities.

\paragraph{Iteration Budget Ablation.}
We finally vary the PSO iteration budget while keeping other settings fixed.
Figure~\ref{fig:iter_ablation} shows that attack performance improves with more iterations and gradually saturates near the default setting, indicating that sufficient exploration is useful whereas excessive optimization brings diminishing returns.
Overall, these ablations verify the joint effectiveness of curved-edge fractal geometry, Fraser-style rendering, and PSO-based exploration in the proposed structured patch space.
\vspace{0.4em}
\begin{center}
\begin{minipage}[t]{0.48\textwidth}
    \vspace{0pt}
    \centering
    \includegraphics[height=2.75cm,keepaspectratio]{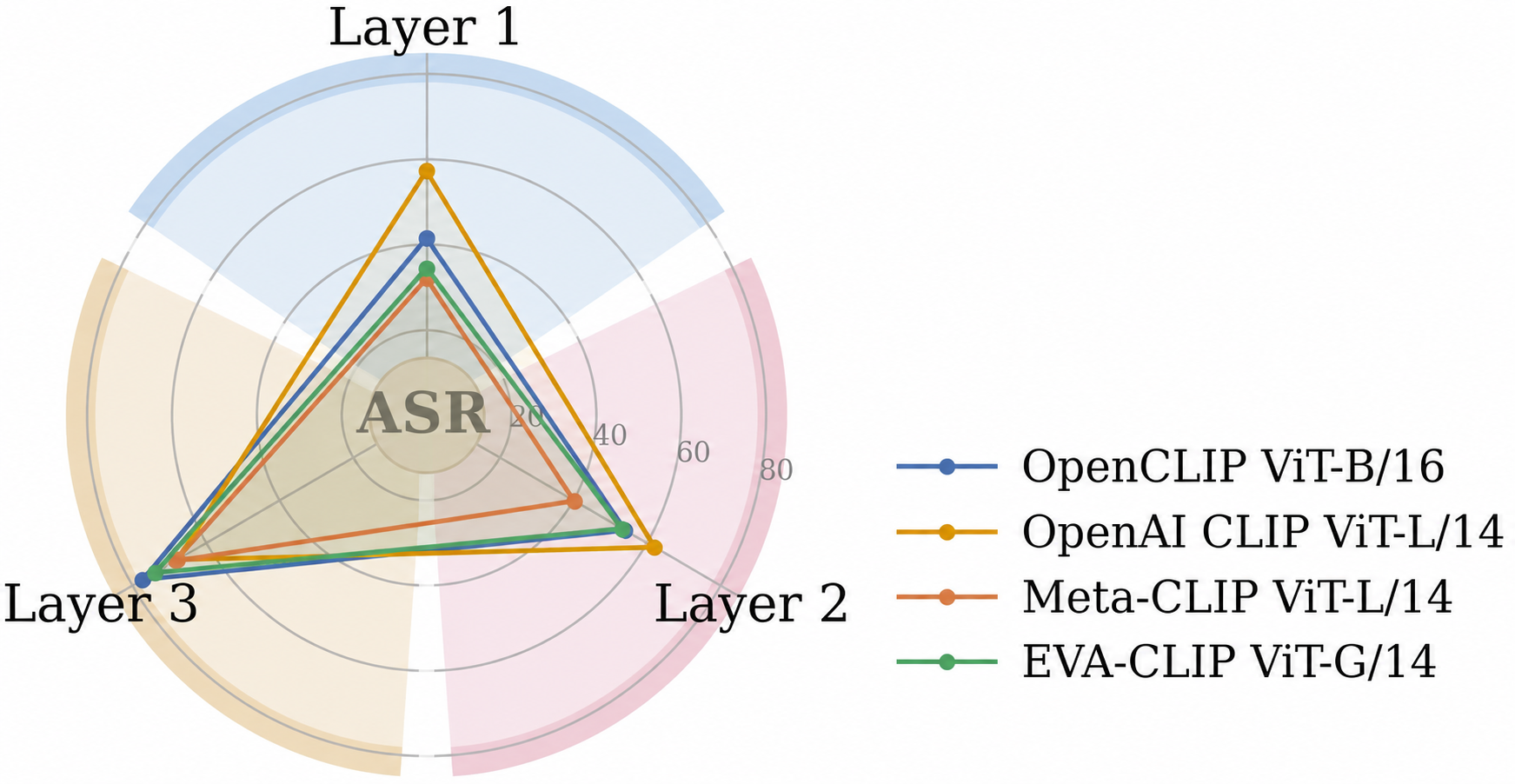}
    \captionof{figure}{Fractal-depth ablation results.}
    \label{fig:depth_ablation}
\end{minipage}
\hfill
\begin{minipage}[t]{0.48\textwidth}
    \vspace{0pt}
    \centering
    \includegraphics[height=2.75cm,keepaspectratio]{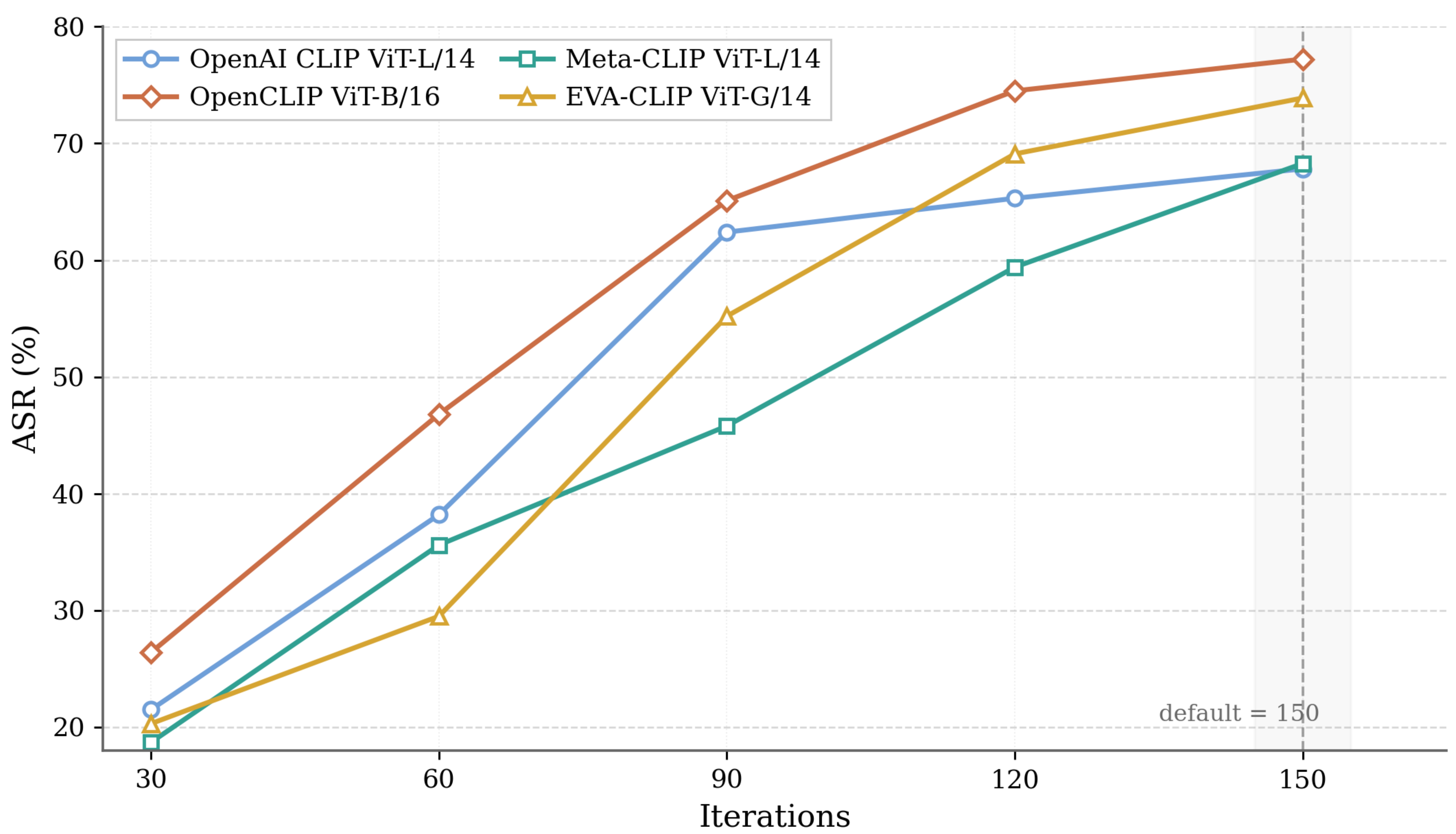}
    \captionof{figure}{Iteration-budget ablation results.}
    \label{fig:iter_ablation}
\end{minipage}
\end{center}
\vspace{-0.6em}
\section{Discussion and Conclusion}

We presented \method{}, a unified geometric adversarial framework for exposing vulnerabilities in VIS--IR VLMs. By combining curved fractal geometry, modality-specific Fraser rendering, and PSO with EOT, \method{} optimizes a compact non-pixel patch space under a strict shared-mask constraint. Experiments on zero-shot classification, image captioning, and VQA show consistent joint VIS--IR attack gains and cross-task transferable semantic degradation, demonstrating that VIS--IR VLMs remain vulnerable to coupled geometric and appearance-level perturbations.

\paragraph{Limitations and Future Work.}
Our work focuses on digital VIS--IR evaluation. In real-world deployment, factors such as printing artifacts, material reflectance, sensor variations, viewpoint changes, and illumination may affect performance. Future work will extend \method{} to physical patch attacks, larger VIS--IR benchmarks, and robustness evaluation under defense scenarios.

{\small
\bibliographystyle{unsrtnat}
\bibliography{r}
}

\clearpage
\appendix

\setlength{\textfloatsep}{12pt plus 2pt minus 2pt} 
\setlength{\floatsep}{12pt plus 2pt minus 2pt}
\setlength{\intextsep}{12pt plus 2pt minus 2pt}

\section{Selected COCO Categories}
\label{sec:appendix_cls}

We evaluate zero-shot image classification on a subset of 30 categories from the COCO dataset, selected to cover diverse semantic groups such as animals, vehicles, and common objects. This subset balances visually distinct and semantically similar classes, providing a compact yet representative testbed for assessing robustness and cross-category generalization of vision--language models under adversarial perturbations. All evaluations are conducted using text prompts without task-specific fine-tuning. The full category list is provided in Table~\ref{tab:coco30}.

\begin{table}[H]
    \centering
    \caption{COCO categories selected for zero-shot image classification.}
    \label{tab:coco30}
    \small
    \setlength{\tabcolsep}{8pt}
    \renewcommand{\arraystretch}{1.1}
    \begin{tabular}{@{} c l @{\hspace{12pt}} c l @{\hspace{12pt}} c l @{}}
        \toprule
        \textbf{ID} & \textbf{Category} &
        \textbf{ID} & \textbf{Category} &
        \textbf{ID} & \textbf{Category} \\
        \midrule
        0  & airplane     & 1  & banana        & 2  & bear \\
        3  & bed          & 4  & bird          & 5  & boat \\
        6  & broccoli     & 7  & bus           & 8  & cake \\
        9  & cell phone   & 10 & clock         & 11 & cow \\
        12 & dog          & 13 & donut         & 14 & elephant \\
        15 & fire hydrant & 16 & horse         & 17 & kite \\
        18 & motorcycle   & 19 & pizza         & 20 & sandwich \\
        21 & teddy bear   & 22 & traffic light & 23 & stop sign \\
        24 & toilet       & 25 & train         & 26 & umbrella \\
        27 & vase         & 28 & zebra         & 29 & sheep \\
        \bottomrule
    \end{tabular}
\end{table}

\section{LLM-as-Judge Prompts for Image Captioning}
\label{sec:appendix_caption_prompt}

\begin{figure}[H]
    \centering
    \includegraphics[width=0.5\linewidth]{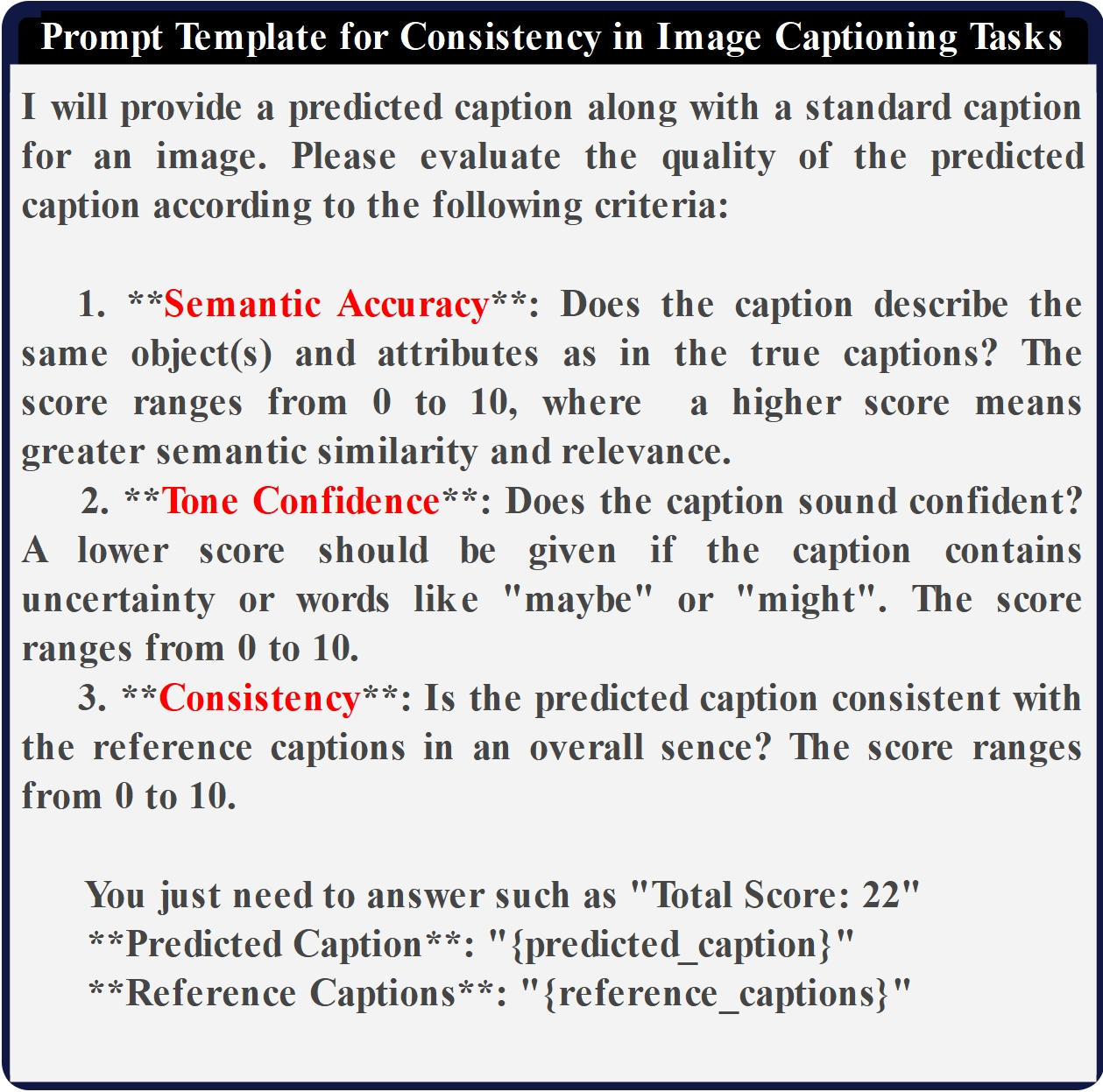}
    \caption{Prompt template for evaluating semantic consistency in image captioning.}
    \label{fig:caption_judge_prompt}
\end{figure}

We adopt an LLM-as-judge framework to evaluate caption consistency under adversarial perturbations. Given an image and its generated caption, the LLM is prompted to assess whether the caption faithfully reflects the visual content, focusing on key objects, attributes, and relationships rather than linguistic fluency. This provides a scalable and reproducible alternative to human evaluation while maintaining strong semantic alignment. The exact prompt template is shown in Figure~\ref{fig:caption_judge_prompt}.

\section{LLM-as-Judge Prompts for Visual Question Answering}
\label{sec:appendix_vqa_prompt}

\begin{figure}[H]
    \centering
    \includegraphics[width=0.5\linewidth]{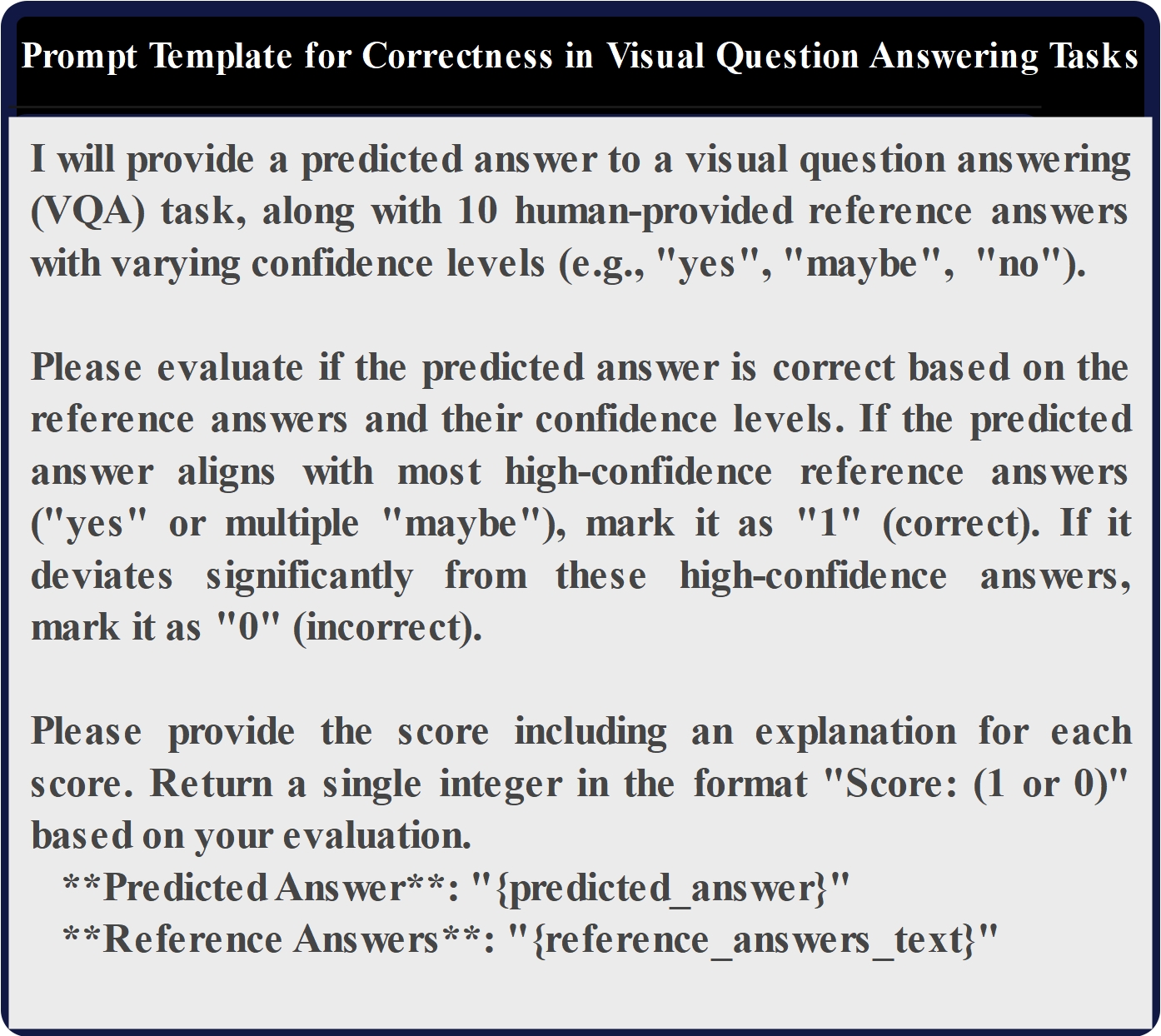}
    \caption{Prompt template for correctness evaluation in visual question answering.}
    \label{fig:vqa_judge_prompt}
\end{figure}

For visual question answering, we extend the LLM-as-judge paradigm to assess answer correctness given an image, a question, and a predicted answer. The LLM is instructed to verify consistency between the answer and the visual evidence, taking into account both question intent and relevant image regions. This unified evaluation protocol enables consistent measurement of adversarial effects across multimodal reasoning tasks. The prompt template used for evaluation is illustrated in Figure~\ref{fig:vqa_judge_prompt}.

\section{CFGPatch Pseudocode}

Algorithm~\ref{alg:cfgpatch} presents the unified optimization procedure of CFGPatch. Given paired visible (VIS) and infrared (IR) images, the method jointly optimizes a shared geometric structure and modality-specific appearance parameters using particle swarm search under EOT. At each iteration, candidate patches are rendered, composited onto both modalities, and evaluated with a joint loss to enforce simultaneous misclassification.

\begin{algorithm}[t]
\caption{CFGPatch: Unified VIS--IR Adversarial Patch Optimization}
\label{alg:cfgpatch}
\begin{algorithmic}[1]

\Require $(I_{\text{vis}}, I_{\text{ir}}), y, N, T, \mathcal{T}, \theta_g, (\theta_{\text{vis}}^a, \theta_{\text{ir}}^a)$
\Ensure $(\theta_g^*, \theta_{\text{vis}}^{a*}, \theta_{\text{ir}}^{a*})$

\State Compute saliency regions and initialize patch pose
\State Initialize swarm $\{x_i^0, v_i^0, b_i^0\}_{i=1}^N$

\For{$t=0$ to $T-1$}
    \For{$i=1$ to $N$}
        \State Construct shared geometry mask $M_g(\theta_g^i)$ and render patches $P_{\text{vis}}, P_{\text{ir}}$
        
        \State $I_{\text{vis}}^{\text{adv}} \gets (1-\alpha_{\text{vis}}M_g)\odot I_{\text{vis}} + \alpha_{\text{vis}}M_g\odot P_{\text{vis}}$
        \State $I_{\text{ir}}^{\text{adv}} \gets (1-\alpha_{\text{ir}}M_g)\odot I_{\text{ir}} + \alpha_{\text{ir}}M_g\odot P_{\text{ir}}$
        
        \State Sample $\tau \sim \mathcal{T}$
        \State $\ell_i \gets \ell_{\text{joint}}(\tau(I_{\text{vis}}^{\text{adv}}), \tau(I_{\text{ir}}^{\text{adv}}))$
        
        \State $b_i^t \gets \arg\min_{x \in \{b_i^t, x_i^t\}} \ell(x)$
    \EndFor

    \State $b_\star^t \gets \arg\min_i \ell(b_i^t)$

    \For{$i=1$ to $N$}
        \State $v_i^{t+1} \gets \omega v_i^t + c_1\xi_1(b_i^t-x_i^t) + c_2\xi_2(b_\star^t-x_i^t)$
        \State $x_i^{t+1} \gets \Pi_B(x_i^t + v_i^{t+1})$
    \EndFor

    \If{$\hat{y}_{\text{vis}}\neq y$ \textbf{and} $\hat{y}_{\text{ir}}\neq y$}
        \State \textbf{break}
    \EndIf
\EndFor

\State \Return $b_\star^T$

\end{algorithmic}
\end{algorithm}

\section{Compute Resources}

All experiments were conducted on a single NVIDIA A100 GPU with 40\,GB VRAM and 83.5\,GB system RAM. Attack optimization and model inference were executed on the GPU, while auxiliary data processing and result aggregation were performed on the CPU. Under the default attack configuration, the shared dual-modality attack was evaluated on 300 paired VIS--IR samples, with a subset of valid samples included in the final evaluation after filtering. Since different vision encoders exhibit varying computational costs, the runtime differed across models, including OpenAI CLIP ViT-L/14, OpenCLIP ViT-B/16, Meta-CLIP ViT-L/14, and EVA-CLIP ViT-G/14. For a single model, the complete evaluation pipeline typically required approximately 11--12 GPU-hours.

\end{document}